\documentclass[preprint]{vgtc}                          

\usepackage{mathptmx}
\usepackage{graphicx}
\usepackage{times}
\usepackage{hyperref}
\usepackage[caption=false,font=footnotesize,labelfont=sf,textfont=sf]{subfig}
\usepackage{float}
\usepackage{multirow}
\usepackage{threeparttable}

\usepackage{color}

\onlineid{0}

\vgtccategory{Research}

\vgtcinsertpkg

\title{Recent Developments and Future Challenges in Medical Mixed Reality}

\author{Long Chen\thanks{e-mail:chenl@bournemouth.ac.uk}\thanks{\href{http://longchen.uk}{http://longchen.uk}}\\ %
        \scriptsize Bournemouth University, UK %
\and Thomas~W~Day\thanks{e-mail:t.day@chester.ac.uk}\\ %
     \scriptsize University of Chester, UK %
\and Wen Tang\thanks{e-mail:wtang@bournemouth.ac.uk}\\ %
     \scriptsize Bournemouth University, UK
\and Nigel~W~John\thanks{e-mail:nigel.john@chester.ac.uk}\\ %
     \scriptsize University of Chester, UK
     }

\abstract{Mixed Reality (MR) is of increasing interest within technology-driven modern medicine but is not yet used in everyday practice. This situation is changing rapidly, however, and this paper explores the emergence of MR technology and the importance of its utility within medical applications. A classification of medical MR has been obtained by applying an unbiased text mining method to a database of 1,403 relevant research papers published over the last two decades. The classification results reveal a taxonomy for the development of medical MR research during this period as well as suggesting future trends. We then use the classification to analyse the technology and applications developed in the last five years. Our objective is to aid researchers to focus on the areas where technology advancements in medical MR are most needed, as well as providing medical practitioners with a useful source of reference.
} 

\CCScatlist{ 
  \CCScat{H.5.1}{Information Interfaces And Presentation}%
{Multimedia Information Systems}{Artificial, augmented, and virtual realities};
  \CCScat{J.3}{Life And Medical Sciences}{Medical information systems}
}

\begin{document}


\firstsection{Introduction}

\maketitle


	In this paper we use the term Mixed Reality (MR) to cover applications of both Augmented Reality (AR) where the virtual augments the real; and Augmented Virtuality where the real augments the virtual, as defined on the Reality-Virtuality Continuum ~\cite{milgram1995augmented}. The use of MR in medicine has long been expected to be a disruptive technology ~\cite{MilgramKishino1994} ~\cite{Azuma1997}, with potential uses in medical education, training, surgical planning, and to guide complex procedures. An early landmark example demonstrated using MR in laparoscopic surgery~\cite{Fuchs1998}. By directly linking patient specific data such as 3D anatomical models with complex surgical scenes, MR can offer a rich source of information to guide intrinsic movements for humans and also for surgical robots.
	
	Technology limitations and costs have been inhibiting factors to the widespread adoption of MR. Nevertheless many different applications and techniques have been published in the last two decades ~\cite{ZhouDuhBillinghurst2008}~\cite{VanKrevelenPoelman2010}. The recent launch of new affordable hardware devices for MR such as Microsoft's HoloLens will be a catalyst for further research. This paper focuses on medical applications, providing a classification of current work and identifying the challenges that must be overcome to narrow the gap between academic research and deployment in clinical practice. Our objective is to aid researchers to focus on the research challenges where technology advancements in medical MR are most needed, as well as providing medical practitioners with a useful source of reference to help them understand how to effectively take advantage of this technology. This is particularly timely with the emergence of many new affordable hardware and software options for MR. 
	
	Section 2 introduces our classification of medical MR. A database of 1,403 relevant publications have been retrieved from the Scopus literature database covering the period 1995 to 2015. A text mining method has been used to identify key topics in medical MR from this database and provides hierarchical classification that can be used for trend analysis and taxonomic review, which is carried out in Section 3. Sections 4 and 5 then review the latest work published in the key topics identified, divided into applications and technologies. Finally, we discuss the current and future research challenges. \par

\section{Classification of Medical MR}

Bibliometric methods are the most common approaches used in identifying research trends by analysing scientific publications\cite{LiDingFengEtAl2009} \cite{HungZhang2012} \cite{VangaSinghVagadiaEtAl2015} \cite{Dey2016}. These methods typically make predictions by measuring certain indicators such as geographical distributions of research institutes and the annual growth of publications, as well as citation counts\cite{GireeshGowdaothers2008}. Usually a manual classification process is carried out ~\cite{Dey2016}, which is inefficient and also can be affected by personal experience. We make use of a generative probabilistic model for text mining  -- Latent Dirichlet Allocation (LDA) ~\cite{BleiNgJordan2003} to automatically classify and generate the categories, achieving an unbiased review process.

Medical MR is an very broad topic that can be viewed as a multidimensional domain with a crossover of many technologies (e.g. camera tracking, visual displays, computer graphics, robotic vision, and computer vision etc.) and applications (e.g. medical training, rehabilitation, intra-operative navigation, guided surgery). The challenge is to identify the research trends across this complex technological and application domain. Our approach is to analyse the relevant related papers retrieved from different periods, whilst introducing a novel method to automatically decompose the overarching topic (medical mixed reality) into relevant sub-topics that can be analysed separately. \par

       \begin{table*}
            \centering
            \caption{Topic Clustering Results from the LDA Model}
            \label{topic_table}
            \begin{tabular}{cc|cc|cc|cc|cc}
            \multicolumn{2}{c|}{\textit{Topic1}}      & \multicolumn{2}{c|}{\textit{Topic 2}}     & \multicolumn{2}{c|}{\textit{Topic 3}}          & \multicolumn{2}{c|}{\textit{Topic 4}}   & \multicolumn{2}{c}{\textit{Topic 5}}    \\
            \multicolumn{2}{c|}{\textbf{``Treatment"}} & \multicolumn{2}{c|}{\textbf{``Education"}} & \multicolumn{2}{c|}{\textbf{``Rehabilitation"}} & \multicolumn{2}{c|}{\textbf{``Surgery"}} & \multicolumn{2}{c}{\textbf{``Training"}} \\
            \textbf{term}       & \textbf{weight}     & \textbf{term}      & \textbf{weight}      & \textbf{term}          & \textbf{weight}       & \textbf{term}     & \textbf{weight}     & \textbf{term}      & \textbf{weight}    \\ \hline
            treatment           & 0.007494            & learning           & 0.01602             & physical               & 0.01383               & surgical          & 0.05450            & training           & 0.03044           \\
            clinical            & 0.007309            & development        & 0.00877             & rehabilitation         & 0.01147              & surgery           & 0.02764            & performance        & 0.01361           \\
            primary             & 0.004333            & education          & 0.00854             & environment            & 0.01112              & surgeon           & 0.01176            & laparoscopic       & 0.01332           \\
            qualitative         & 0.003793            & potential          & 0.00812             & game                   & 0.00837              & invasive          & 0.01167            & skills             & 0.01208           \\
            focus               & 0.004165            & different          & 0.00793             & therapy                & 0.00729              & minimally         & 0.01148            & simulator          & 0.01198              \\ \\

            \multicolumn{2}{c|}{\textit{Topic6}}        & \multicolumn{2}{c|}{\textit{Topic 7}}  & \multicolumn{2}{c|}{\textit{Topic 8}}   & \multicolumn{2}{c|}{\textit{Topic 9}}        & \multicolumn{2}{c}{\textit{Topic 10}}   \\
            \multicolumn{2}{c|}{\textbf{``Interaction"}} & \multicolumn{2}{c|}{\textbf{``Mobile"}} & \multicolumn{2}{c|}{\textbf{``Display"}} & \multicolumn{2}{c|}{\textbf{``Registration"}} & \multicolumn{2}{c}{\textbf{``Tracking"}} \\
            \textbf{term}        & \textbf{weight}      & \textbf{term}     & \textbf{weight}    & \textbf{term}     & \textbf{weight}     & \textbf{term}        & \textbf{weight}       & \textbf{term}     & \textbf{weight}     \\ \hline
            human                & 0.019901             & software          & 0.01684           & visualization     & 0.03260            & registration         & 0.01417              & tracking          & 0.02418            \\
            interaction          & 0.014849             & mobile            & 0.01556           & data              & 0.03177            & segmentation         & 0.00942              & accuracy          & 0.01584            \\
            haptic               & 0.01439              & support           & 0.00905           & display           & 0.00984            & accurate             & 0.00765              & camera            & 0.01454            \\
            feedback             & 0.013308             & online            & 0.00874           & navigation        & 0.01278            & deformation          & 0.00762              & target            & 0.01347            \\
            interface            & 0.009382             & social            & 0.00835           & planning          & 0.01225            & motion               & 0.00754              & registration      & 0.01186           
            \end{tabular}
        \end{table*}
        
	\subsection{Data Source}
	The source database used in this analysis is Scopus, which contains the largest abstract and citation database of peer-reviewed literature obtained from more than 5,000 international publishers\cite{Content_scopus}. Scopus contains articles published after 1995 \cite{facts_wok}, therefore, encompassing the main period of growth in MR, and helps both in keyword searching and citation analysis.
	
	\subsection{Selection Criteria}
		The regular expression ``(mixed OR augmented) reality medic*'' is used to retrieve all articles related to augmented reality and mixed reality in medicine, capturing ``augmented reality medicine'', ``augmented reality medical'', and ``mixed reality medicine'', ``mixed reality medical''. A total of 1,425 articles were retrieved within the 21 year period from 1995 to 2015, of which 1,403 abstracts were accessed. We initially categorised these articles into seven chronological periods, one for every three years. Abstracts of these articles are then used to generate topics and for trend analysis, as they provide more comprehensive information about an article than its title and keywords alone. The whole selection process is carried out automatically with no manual intervention.
		
	        \subsection{Text Mining}
		To identify the key topics discussed in a large number of articles, we employ the Latent Dirichlet Allocation (LDA)\cite{BleiNgJordan2003} method to automatically interpret and cluster words and documents into different topics. This \textit{text mining method} has been widely used in recommendation systems such as web search engines and advertising applications. LDA is a generative probabilistic model of a corpus. It regards documents ($d$) as random mixtures over latent topics ($t$), $p(t|d)$, where every topic is characterized by a distribution over words ($w$), $p(w|t)$. The method uses the following formula: 
		\begin{equation}
		    \label{lda_formula}
		    p(w|d) = p(w|t) * p(t|d)
		\end{equation}
		where $p(w|d)$ represent the probability of a certain word in a certain document under a certain topic. Word-topic distribution $p(w|t)$ and topic-document distribution $p(t|d)$ are randomly selected and then LDA iteratively updates and estimates probabilities until the system convergences. As a result, we identify the relationships amongst documents, topics and words. 
                We input all of the downloaded abstracts into the LDA model and tested a range of parameters to use with it. We empirically derived the value of ten topics as  optimal for the best encapsulation of the field. 
                
    \subsection{Topic Generation}
                
                Table \ref{topic_table} summarizes the output showing the ten topics identified with the associated term and weight distributions after convergence. We  manually assign one word (shown in quotation marks) that was the best representation of each topic. The general methodology uses the weighting as the primary selection parameter but also takes into account the descriptive keyword for that topic. Topics 1, 5, 9 and 10 just use the keyword with the highest weighting. For Topic 2, although ``education" did not have the highest weighting, we consider it to be a more representative term. For Topic 3, ``physical" is a sub category of ``rehabilitation'' and so we use the latter as the more generic term. In Topic 4, ``surgical" and ``surgery" are fundamentally the same. For Topic 6, ``interaction" is the most representative keyword, and the same principle applies to Topics 7 and 8.

		\begin{figure}[htb]
    		\centering
    		\includegraphics[width=3in]{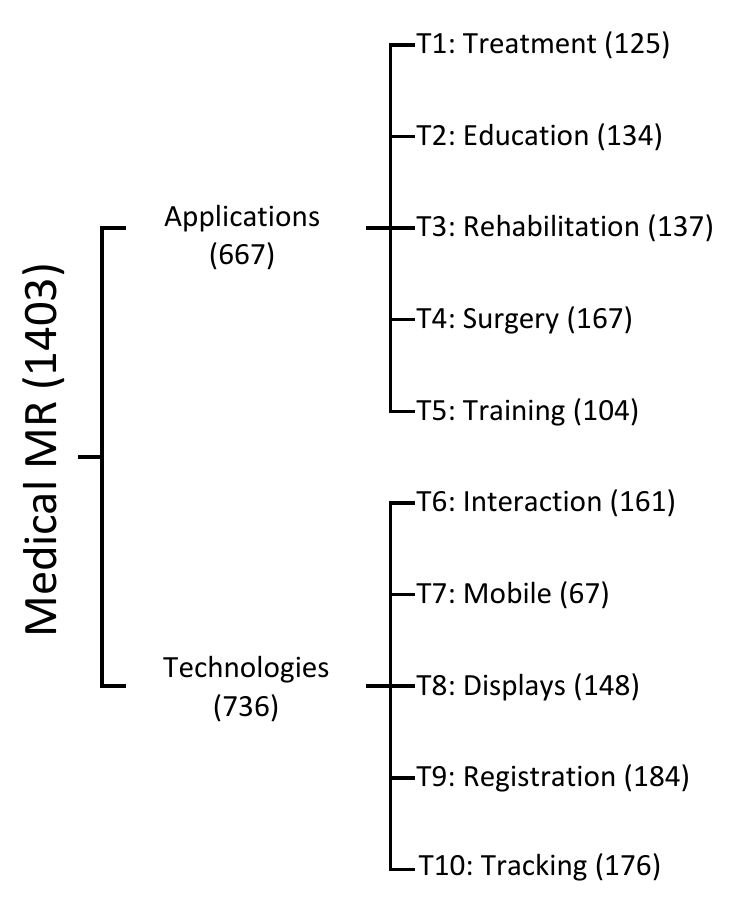}
    		\caption{Hierarchical taxonomy of Medical MR}
    		\label{topic_map}
	    \end{figure}
		Figure \ref{topic_map} represents a hierarchical taxonomy of the results. The overarching ``Medical MR`` topic with 1,403 articles has been divided into two main sub-categories: applications and technologies, with 667 and 736 articles respectively. Within applications, the surgical topic has the largest number of articles (167), followed by rehabilitation (137), education (134), treatment (125) and training (104). Within technologies, registration is the most discussed topic (184 articles), followed by tracking (176), interaction (161), displays (148) and mobile technologies (67).

	 \section{Trend Analysis}
	   
\begin{figure}[htb]
    		\centering
    		\subfloat[]{\includegraphics[width=3.5in]{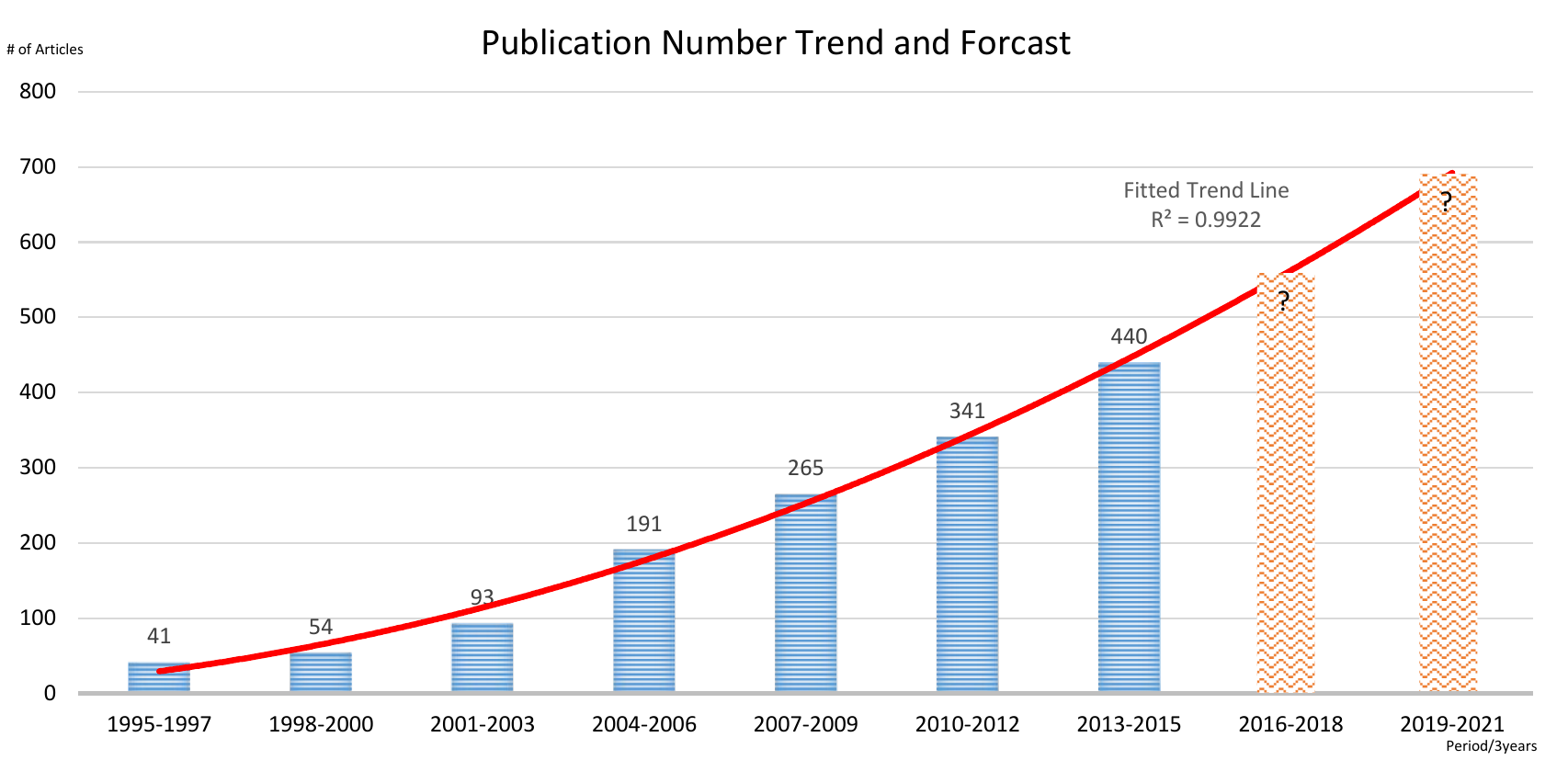}}\\
        	\subfloat[]{\includegraphics[width=3.5in]{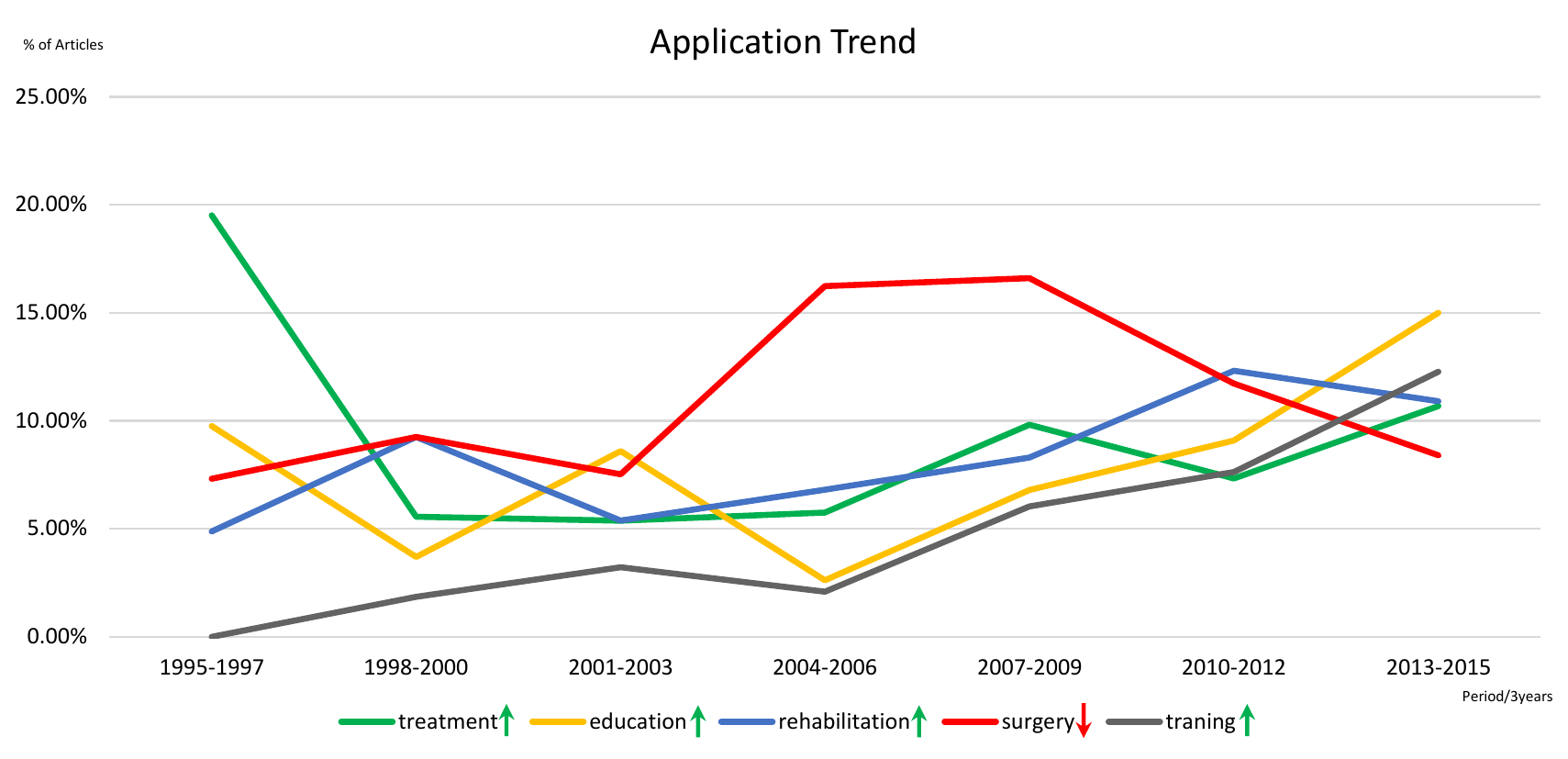}}\\
        	        	\subfloat[]{\includegraphics[width=3.5in]{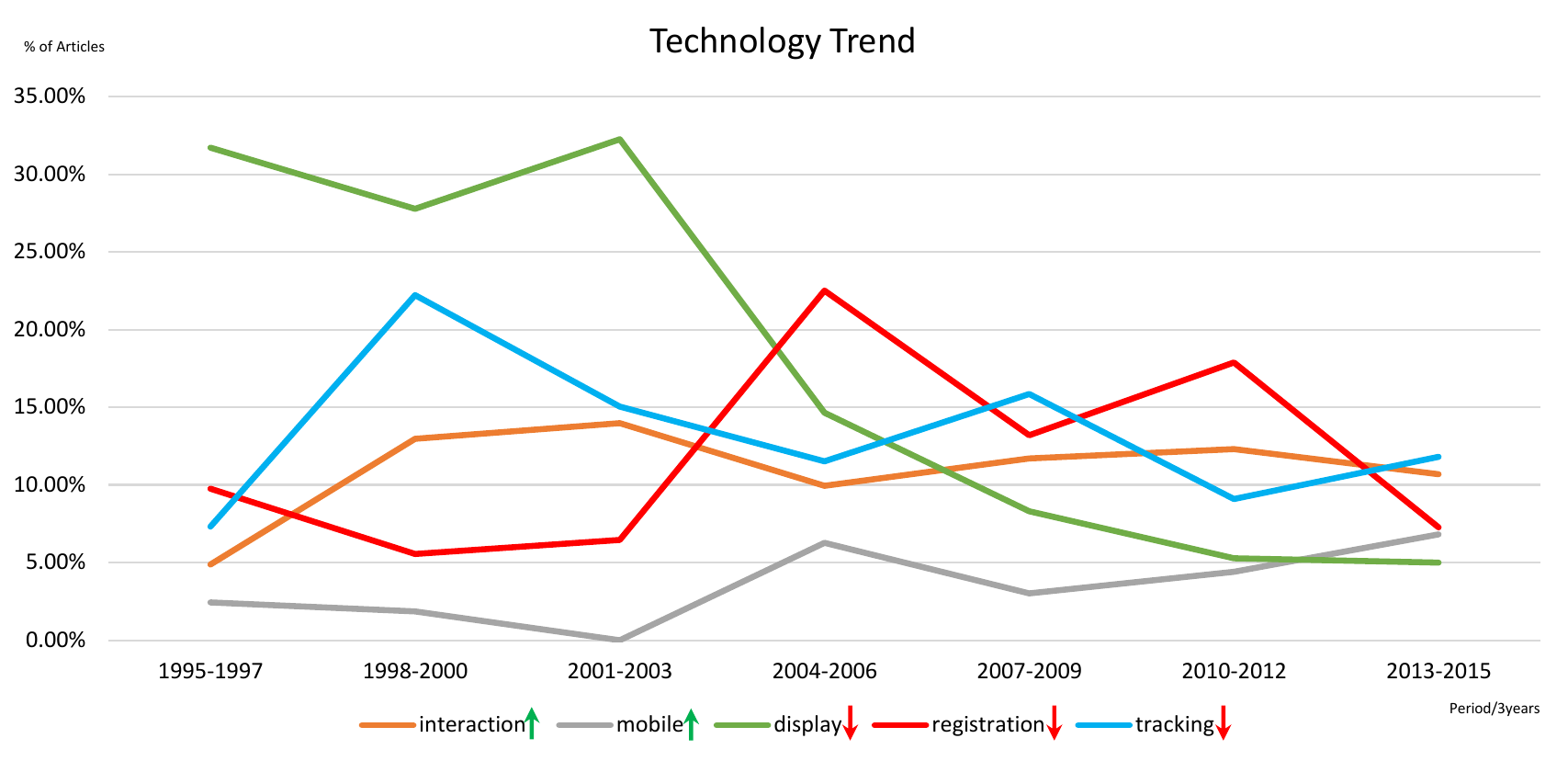}}\\
        	\caption{Trend analysis: (a) Publication Trends. (b) Application Trends. (c) Technology Trends.}
        	\label{trend_fig}
    	\end{figure}

                Each of the ten topics of interest identified by the LDA model has a list of articles associated with them. An article matrix was constructed based on the topic and the attributes for the seven chronological periods being analysed.  Figure \ref{trend_fig} summarizes the trends identified subdivided into three year periods (1995-97, 1998-2000, 2001-03, 2004-06, 2007-09, 2010-12, and 2013-15). Figure \ref{trend_fig}(a) plots the total number of publications over the seven periods. The number of publications related to MR in medicine has increased more than 100 times from only 41 publications in 1995-1997 to 440 publications in 2013-2015. In the early 21st century (periods 2001-2003 and 2004-2006), the number of publications of MR in medicine more than doubled from 93 to 191, coinciding with the rapid development of many enabling technologies such as marker-based tracking techniques\cite{KatoBillinghurst1999} and advances in Head Mounted Display (HMD) technology ~\cite{RollandFuchs2000} and mobile AR devices ~\cite{olsson2011online}.\par

    	Based on the observed growth pattern between 1995 and 2015, a trend line has been produced using a quadratic polynomial with a high coefficient of determination ($R^2 = 0.9935$). Extrapolating the trend line forecasts that in the three year periods (2016 to 2018, and 2019 to 2021), the number of scientific papers on the topic of MR in medicine will be accelerated, reaching around 550 and 700, respectively. The following section looks in more detail at the topic trends and then we analyse the research trends in each area.

	\subsection{Applications Trends}

        There are a growing number of medical application areas exploring the use of MR. Fig. \ref{trend_fig}(b) plots the percentage of articles published for the five most popular application categories: \textit{patient treatment, medical and patient education, rehabilitation, surgery,} and \textit{procedures training}:
    	
    	\begin{itemize}

    	\item Patient treatment was the most targeted application of MR in the earlier period with almost 20\% of published articles. It remains a constant topic of interest with around 10\% of articles continuing to investigate this topic. The fall in percentage is mostly due to the parallel rise in interest in the other medical MR categories. Education and rehabilitation topics have both fluctuated but remain close to 10\% of articles published.
    	
        \item A surge of interest in surgical applications can be observed between 2004 and 2009 when 16\% of all articles published on medical MR addressed this topic. However, the comparative dominance of surgical applications has dropped off as activity within other categories has increased.
    	
        \item Training applications in medical MR first emerged between 1998-2000. Interest in this topic has grown steadily and is also now at a similar level of interest as the other topics. Together with education, continuation of the current trends suggest that these two topics will be the most popular in the next few years. These are areas where there is no direct involvement with patients and so ethical approval may be easier to gain.\par
        \end{itemize}
        
	\subsection{Technologies Trends}
    	Within the ten topics generated by the LDA model, five key technologies have been identified:  \textit{interaction, mobile, display, registration} and \textit{tracking} (the percentage of articles that refer to these technologies is plotted in Fig. \ref{trend_fig}(c)):
    	
    \begin{itemize}
    	\item Real time interaction is a crucial component of any MR application in medicine especially when interactions with patient is involved. The percentage of articles that discuss interaction in the context of medical MR increased steadily from 5\% in 1995-1997 to 10\% in 2013-2015. \par
    	
    	\item The use of mobile technologies is an emerging trend, which has been increased from 0\% to 7\% of articles across the seven periods. The main surge so far was from 2004-2006, when the advances of micro-electronics technology firstly enabled mobile devices to run fast enough to support AR applications. The use of mobile technologies has been more or less constant from that point onwards. Smartphones and tablets can significantly increase the mobility and user experience as you are not tethered to a computer, or even in one fixed place as was the case with Sutherland's \cite{Sutherland1968} first AR prototype in the 1960s. \par
    	
    	\item Innovations in the use of display technologies was the most discussed MR topic in the early part of the time-line. However, there has been a subsequent dramatic drop in such articles, falling from  33\% of articles to only 5\%.  This may indicate the maturity of the display technologies currently in use. The Microsoft HoloLens and other new devices are expected to disrupt this trend, however.\par
    	
    	\item Tracking and registration are important enabling components for MR. They directly impact on the usability and performance of the system. These areas continue to be explored and are yet to be mature enough for complex scenarios, as reflected by the steady percentage (around 10\%) of articles on tracking and registration technology from 1995 to 2015.\par
\end{itemize}

        In the next two sections we summarise the latest research in medical MR using the classification scheme identified above. We restrict our analysis to publications in the last five years, citing recent survey papers wherever possible.

\section{Applications}

\subsection{Treatment}

There is no doubt that the use of MR can assist in a variety of patient treatment scenarios. Radiotherapy treatment is one example and Cosentino \textit{et al.} ~\cite{cosentino2014an} have provided a recent review. Most of the studies identified in this specialty area indicate that MR in radiotherapy still needs to address limitations around patient discomfort, ease of use and sensible selection and accuracy of the information to be displayed, although the required accuracy for most advanced radiotherapy techniques is of the same order of magnitude as that which is already being achieved with MR. 

\begin{figure}[!htb]
	\centering
	\includegraphics[width=3.3in]{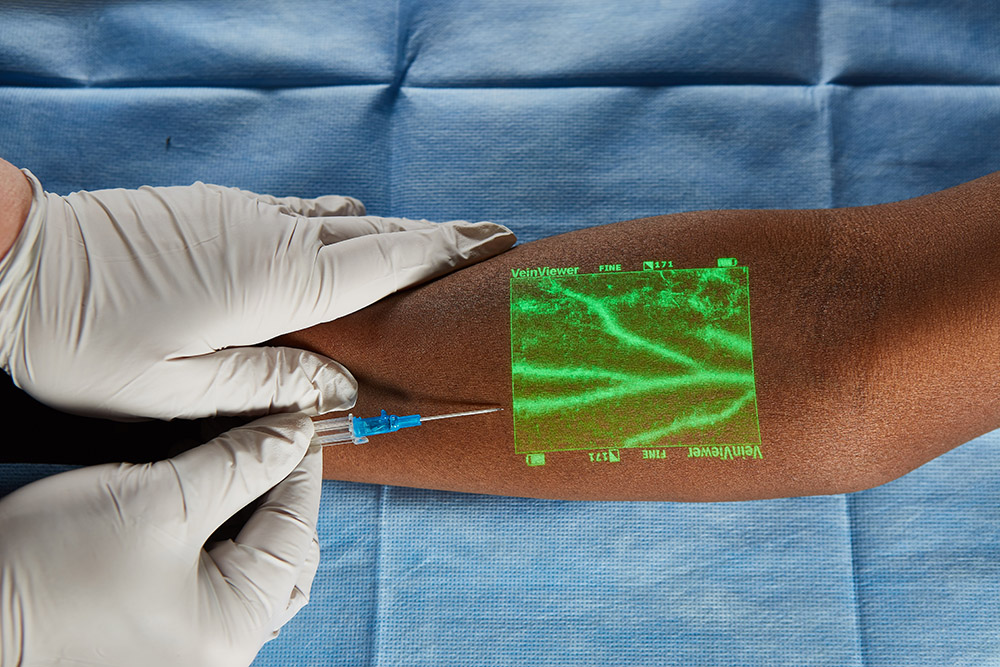}
	\caption{The VeinViewer\textsuperscript{\textregistered} Vision system projects the patient's vascular network onto the skin to help with needle insertion. Image courtesy of Christie Medical Holdings.}
	\label{treatment_fig}
\end{figure}

Treatment guidance in general can be expected to benefit from MR. One of the only commercial systems currently exploiting AR in a hospital setting is VeinViewer Vision (Christie Medical Holdings, Inc, Memphis, TN), a system to assist with vascular access procedures. It projects near-infrared light onto the skin of a patient, which is absorbed by blood and reflected by surrounding tissue. This information is captured in real time and allows an image of the vascular network to be projected back onto the skin providing an accurate map of patient's blood vessel pattern - see Figure \ref{treatment_fig}. This makes many needle puncture procedures such as gaining IV access easier and to be performed successfully ~\cite{kim2012efficacy}. However, another study with skilled nurses found that the success of first attempts with VeinViewer actually worsened ~\cite{szmuk2013veinviewer}, which highlight the need for further clinical validation. The ability to identify targets for needle puncture or insertion of other clinical tools  is an active area of development.

\subsection{Rehabilitation}
Patient rehabilitation is a broad term that covers many different types of therapies to assist patients in recovering from a debilitating mental or physical ailment. Many studies report that MR provides added motivation for stroke patients and those recovering from injuries who often get bored by the repetition of the exercises they need to complete (e.g. ~\cite{aung2014novel}, ~\cite{hossain2015ar}, ~\cite{vidrios2015development}). Often solutions are based on affordable technologies that can be deployed in patients homes. MR systems can also be used to monitor that the patient is performing exercises correctly (e.g. ~\cite{tan2014short}).

Neurological rehabilitation is another area where MR is being applied. Caraiman \textit{et al.} ~\cite{caraiman2015architectural} present a neuromotor rehabilitation system that provides augmented feedback as part of a learning process for the patient, helping the brain to create new neural pathways to adapt. 

Rehabilitation is probably the most mature of the medical application areas currently using MR, and will continue to flourish as more home use deployment becomes possible.
\subsection{Surgery}
Surgical guidance can benefit from MR by providing information from medical scan images to a surgeon in a convenient and intuitive manner. Kerstan-Oertel \textit{et al.} ~\cite{kersten2013state} provide a useful review of mixed reality image guided surgery and highlight four key issues: the need to choose appropriate information for the augmented visualization; the use of appropriate visualization processing techniques; addressing the user interface; and evaluation/validation of the system. These remain important areas of research. For example, a recent study using 50 experienced otolaryngology surgeons found that those using an AR display were less able to detect unexpected findings (such as a foreign body) than those using the standard endoscopic monitor with a submonitor for augmented information ~\cite{dixon2014inattentional}. Human computer interaction (HCI) therefore remains an unsolved problem for this and many other aspects of medical MR.

Most surgical specialties areas are currently experimenting with the use of MR. One of the most active areas is within neurosurgery and Meola \textit{et al.} ~\cite{meola2016augmented} provide a systematic review. They report that AR is a versatile tool for minimally invasive procedures for a wide range of neurological diseases and can improve on the current generation of neuronavigation systems. However, more prospective randomized studies are needed, such as ~\cite{cabrilo2014augmented} and ~\cite{inoue2013preliminary}. There is still a need for further technological development to improve the viability of AR in neurosurgery, and new imaging techniques should be exploited.

Minimally invasive surgical (MIS) procedures is another growth area for the use of MR. Such procedures may be within the abdominal or pelvic cavities (laparoscopy); or the thoracic or chest cavity (thoracoscopy). They are typically performed far from the target location through small incisions (usually 0.5-1.5 cm) elsewhere in the body. The surgeons field of view (FOV) is  limited to the endoscope's camera view and his/her's depth perception is usually dramatically reduced. Nicolau \textit{et al.} ~\cite{NicolauSolerMutterEtAl2011} discus the principles of AR in the context of laparoscopic surgical oncology and the benefits and limitations for clinical use. Using patient images from medical scanners, AR can improve the visual capabilities of the surgeon and compensate for their otherwise restricted field of view. Two main processes are useful in AR: 3D visualization of the patient data; and registration of the visualization onto the patient. The dynamic tracking of organ motion and deformation has been identified as a limitation of AR in MIS ~\cite{nakamoto2012current}, however. Vemuri \textit{et al.} ~\cite{vemuri2012deformable} have proposed a deformable 3D model architecture that has been tested in twelve surgical procedures with positive feedback from surgeons. A good overview of techniques that can be used to improve depth perception is also given by Wang \textit{et al.} \cite{wang2016visualization}. They compared several techniques qualitatively: transparent overlay, virtual window, random-dot mask, ghosting, and depth-aware ghosting; but have not yet obtained quantitative results from a study with subject experts. 

Recent uses of MR in surgery have also been reported for: hepatic surgery 
~\cite{haouchine2013image} ~\cite{haouchine2013deformation} ~\cite{gavaghan2011portable}; eye surgery ~\cite{ong2015augmented};   oral and maxillofacial surgery ~\cite{WangSuenagaHoshiEtAl2014}  ~\cite{badiali2014augmented}; nephrectomy (kidney removal) ~\cite{edgcumbe2016augmented}; and transpyloric feeding tube placement \cite{bano2016augmented}. It is apparent that all surgical specialties can gain value from the use of MR. However, many of the applications reported are yet to be validated on patients in a clinical setting, which is time consuming task but a vital next step.

Industry adoption of the technology will also be key and is starting to occur. In early 2017, Philips announced their first augmented reality system designed for image-guided spine, cranial and trauma surgery. It uses high-resolution optical cameras mounted on the flat panel X-ray detector of a C-arm to image the surface of the patient. It then combines the external view captured by the cameras and the internal 3D view of the patient acquired by the X-ray system to construct a MR view of the patient’s external and internal anatomy. Clinical cases using this system have been successfully reported \cite{racadio2016augmented} \cite{elmi2016surgical}. It can be expected that the support of major medical equipment manufactures will greatly accelerate the acceptance and adoption of mixed reality in the operating room.

\subsection{Training}
\begin{figure}[htb]
		\centering
		\includegraphics[width=3in]{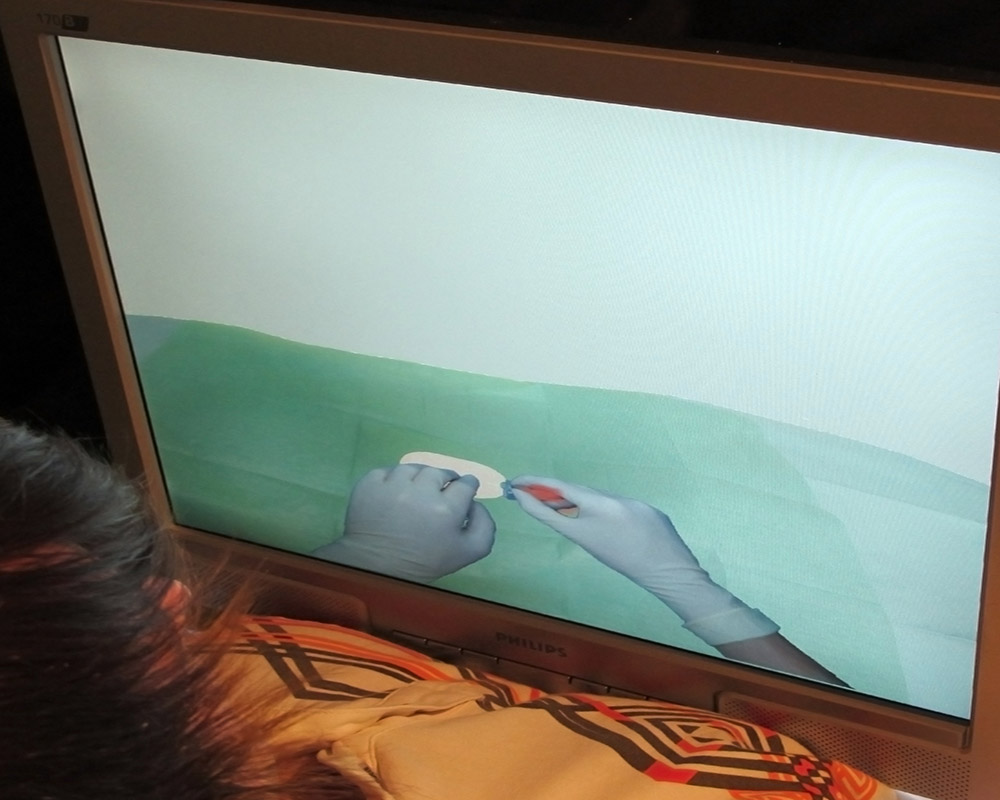}
		\caption{Training simulator for needle puncture guided by palpation. An example of augmented virtuality where the scene of the virtual patient is augmented with the trainees real hands. }
		\label{palpsim}
	\end{figure}

The training of medical/surgical procedures is often a complex task requiring high levels of perceptual, cognitive, and sensorimotor skills. Training on patients has been the accepted practice for centuries, but today the use of intelligent mannequins and/or virtual reality simulators have become a widely accepted alternative. MR can also offer added value to the use of these tools in the training curricula and are starting to appear. For example, the PalpSim system ~\cite{ColesJohnGouldEtAl2011} is an augmented virtuality system where the trainee can see their own hands palpating the virtual patient and holding a virtual needle - see Figure \ref{palpsim}. This was an important factor for the interventional radiologists involved in the validation of the system as it contributed to the fidelity of the training experience. PalpSim and other examples (e.g. ~\cite{yeo2011effect} ~\cite{strickland2011development} ~\cite{TangWGHJ12} ~\cite{abhari2015training} ~\cite{cheng2014augmented}) demonstrate that MR can be used in training to improve the accuracy of carrying out a task.

\subsection{Education}

As well as procedures training, medical education will encompass basic knowledge acquisition such as learning human anatomy, the operation of equipment, communications skills with patients, and much more. 

Recent reviews of MR and AR in medical education ( ~\cite{monkman2015see}, ~\cite{zhu2014augmented}) highlight the traditional reliance on workplace learning to master complex skills. However, safety issues, cost implications and didactics sometimes mean that training in a real-life context is not always possible. Kamphius \textit{et al.} ~\cite{kamphuis2014augmented} discuss the use of AR via three case studies: visualizing human anatomy; visualizing 3D lung dynamics; and training laparoscopy skills ~\cite{kamphuis2014augmented}. The latest work published in these areas is summarised and the important research questions that need to be addressed are identified as:
\begin{itemize}
\item To what extent does an AR training system use a representative context, task, and behaviour compared with the real world? 
\item What learning effects does the AR training system generate?
\item What factors influence the implementation of an AR training system in a curriculum and how does that affect learning?
\item To what extent do learning results acquired with the AR training system transfer to the professional context?
\end{itemize}
Currently most of the emphasis on AR in education is not on these questions but on the development, usability and initial implementation of the AR system. This indicates a significant knowledge gap in AR for medical eduction and will become more important as  medical schools move away from cadavaeric dissection and embrace digital alternatives. 

\section{Technologies}
All of the applications discussed in the previous section rely on a core set of technologies to enable a MR environment, enhanced with application specific tools and devices.
\subsection{Interaction}
Perhaps the most natural interaction is to use our own hands to interact with virtual objects in the augmented environment. Computer vision based tracking using markers attached to fingertips has been shown to correspond to virtual fingers with physical properties such as friction, density, surface, volume and collision detection for grasping and lifting interactions ~\cite{BoonbrahmKaewrat2014}. Wireless instrumented data gloves can also capture finger movements and have been used to perform zoom and rotation tasks, select 3D medical images, and even typing on a floating virtual keyboard in MR ~\cite{DeMarsicoLevialdiNappiEtAl2014}. An alternative is to use infra red (IR) emitters and LEDs to generate optical patterns that can be detected by IR cameras. This type of technology offers touchless interactions suitable for medical applications ~\cite{SousaSilvaFormicoRodrigues2015}. 

 The sense of touch also provides important cues in medical procedures. One advantage of MR over virtual reality applications is that the physical world is still available and can be touched. Any haptic feedback (tactile or force feedback) from virtual objects, however, must make use of specialized hardware ~\cite{ColesMeglanJohn2011}. It may also be useful to augment haptics onto real objects (Haptic Augmented Reality). Examples include simulating the feel of a tumour inside a silicone model~\cite{JeonChoiHarders2012}, and to simulate breast cancer palpation \cite{JeonKnoerleinHardersEtAl2010}.
 
 The deployment of interaction technologies with a clinical setting such as an operating theatre is a particular challenge. The equipment used must not be obtrusive to the procedure being performed, and often has to be sterilised.
   
\subsection{Mobile AR}

        \begin{table*}[htb]
            \centering
            \caption{Mobile Medical MR Applications}
            \label{mobile_ar_app}
            \begin{threeparttable}
            \begin{tabular}{llll}
            \hline
            \textbf{Article}                                                          & \textbf{Purpose}                              & \textbf{SDK}                               & \textbf{Device}  \\ \hline
            \textbf{Andersen \textit{et al} (2016) \cite{AndersenEtAl2016}}       & Surgical Telementoring                        & OpenCV                                    & Project Tango \& Samsung Tablet\\ 
            \textbf{Rantakari \textit{et al} (2015) \cite{RantakariColleyHaekkilae2015}}       & Personal Health Poster                        & Vuforia                                    & Samsung Galaxy S5\\ 
            \textbf{Kilgus \textit{et al} (2015) \cite{KilgusHeimHaaseEtAl2015}}               & Forensic Pathological Autospy                 & MITK\tnote{*}        & Apple iPad 2      \\ 
            \textbf{Soeiro \textit{et al} (2015) \cite{SoeiroClaudioCarmoEtAl2015}}            & Brain Visulization                            & Metaio                                     & Samsung Galaxy S4  \\ 
            \textbf{Juanes \textit{et al} (2014) \cite{JuanesHernandezRuisotoEtAl2014}}        & Human Anatomy Education                       & Vuforia                                    & Apple iPad       \\ 
            \textbf{Kramers \textit{et al} (2014) \cite{KramersArmstrongBakhshmandEtAl2014}}   & Neurosurgical Guidance                        & Vuforia                                    & HTC Smartphone    \\ 
            \textbf{mARble\textsuperscript{\textregistered} (2014) \cite{NollHaeussermannJanEtAl2014}}  & Dermatology Education       & Not Mentioned                              & Apple iPhone 4      \\ 
            \textbf{Mobile RehApp\textsuperscript{TM} (2014) \cite{GarciaNavarro2014}}& Ankle Sprain Rehabilitation                   & Vuforia                                    & Apple iPad          \\ 
            \textbf{Virag \textit{et al} (2014) \cite{ViragStoicu-TivadarAmaricai2014}}        & Medical Image Visulization                    & JSARToolKit                                & Any device with browser\\ 
            \textbf{Grandi \textit{et al} (2014) \cite{GrandiMacielDebarbaEtAl2014}}           & Surgery Planning                              & Vuforia                                    & Apple iPad 3        \\ 
            \textbf{Debarba \textit{et al} (2012) \cite{DebarbaGrandiMacielEtAl2012}}          & Hepatectomy Planning                          & ARToolkit                                  & Apple iPod Touch     \\ 
            \textbf{Ubi-REHAB (2011) \cite{Choi2011}}                                 & Stroke Rehabilitation                         & Not Mentioned                              & Android Smartphone  \\ \hline          
            \end{tabular}

             \begin{tablenotes}
                \item[*] Medical Imaging Interaction Toolkit
            \end{tablenotes}
            \end{threeparttable}
        \end{table*}
        
The rapid development of smartphones and tablets with high quality in-built cameras are providing new opportunities for MR, particularly affordable AR applications. Software Development Kits (SDKs) such as ARToolKit \cite{ARToolKit2016} and Vuforia \cite{PTC2016} are enabling more and more applications to appear. Mobile AR is expected to play a major role in medical/patient education and rehabilitation applications where accuracy of tracking is not critical. Table \ref{mobile_ar_app} provides a summary of mobile medical AR apps currently available. Most use marker-based tracking that rely on the position and focus of the markers and will not work in poor lighting conditions. The display of patient specific data and 3D anatomical models will also be restricted on very small displays. Although some mobile AR applications are used in surgical planning and guidance, these are currently only prototypes built to demonstrate feasibility of using AR, and yet to gain regulatory approval.

\subsection{Displays}
\begin{figure}[!htb]
        		\centering
        		\includegraphics[width=3.3in]{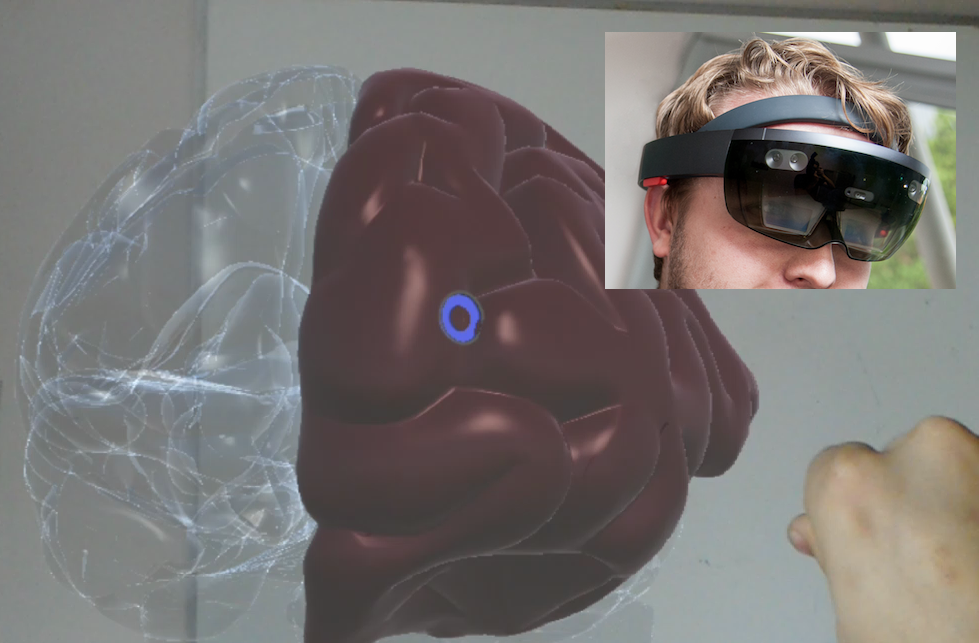}
            	\caption{A MR brain anatomy lesson using the Microsoft HoloLens (see insert), an optical HMD.}
            	\label{hololens}
        	\end{figure}

The HMD has been used in many medical MR applications - see Table \ref{HMD_based_application}. A video see-through HMD captures video via a mono- or stereo-camera, and then overlays computer-generated content onto the real world video. The optical see-through HMD allows users to directly see the real world while virtual objects are displayed through a transparent lens. Users will receive light from both the real world and the transparent lens and form a composite view of real and virtual object. The Google Glass and Microsoft HoloLens (see Figure \ref{hololens}) are recent examples of an optical see-through HMDs. Hybrid solutions that use optical see-through technology for display and video technology for object tracking may play a key role in future developments.

\begin{table*}[htb]
                \centering
                \caption{HMD-based Medical MR Applications}
                \label{HMD_based_application}
                
                \tabcolsep 0.02in 
                \begin{threeparttable}
                \begin{tabular}{lllll}
                \hline
                \textbf{Article}                                             & \textbf{Purpose}                                 &\textbf{HMD Type} & \textbf{HMD Device}                & \textbf{Tracking} \\ \hline
                \textbf{Meng \textit{et al} (2015) \cite{MengShahzadSaadEtAl2015}}           & Veins Localization                        & Optical           & Vuzix STAR 1200XL(modified)        & Manually Aligned  \\ 
                \textbf{Chang \textit{et al} (2015) \cite{ChangHsiehHuangEtAl2015}}          & Remote Surgical Assistance                & Video             & VUNIX iWear VR920                  & Optical Tracker + KLT\tnote{*}         \\ 
                \textbf{Hsieh \textit{et al} (2015) \cite{HsiehLee2015}}                     & Head CT Visulization                      & Video             & Vuzix Wrap 1200DXAR                & KLT + ICP\tnote{**}         \\ 
                \textbf{Wang \textit{et al} (2015) \cite{WangWangLeongEtAl2015}}              & Screw Placement Navigation               & Optical           & NVIS nVisor ST60                   & Optical Tracker   \\ 
                \textbf{Vigh \textit{et al} (2014) \cite{VighMuellerRistowEtAl2014}}          & Oral Implantology                        & Video             & NVIS nVisor SX60                   & Optical Tracker  \\ 
                \textbf{Hu \textit{et al} (2013) \cite{HuWangSong2013}}                      & Surgery Guidance                          & Video             & eMagin Z800 3D Visor               & Marker            \\ 
                \textbf{Abe \textit{et al} (2013) \cite{AbeSatoKatoEtAl2013}}                 & Percutaneous Vertebroplasty              & Video             & HMD by Epson (Model Unknow)        & Marker            \\ 
                \textbf{Azimi \textit{et al} (2012) \cite{AzimiDoswellKazanzides2012}}       & Navigation in Neurosurgery                & Optical           & Goggles by Juxtopia               & Marker            \\ 
                \textbf{Blum \textit{et al} (2012) \cite{BlumStauderEulerEtAl2012}}           & Anatomy Visulization                     & Video             & Not Mentioned                      & Gaze-tracker      \\ 
                \textbf{Toshiaki (2010)  \cite{Tanaka2010}}                         & Cognitive Disorder Rehabilitation         & Video             & Canon GT270                           & No Tracking       \\ 
                \textbf{Wieczorek \textit{et al} (2010) \cite{WieczorekAichertKutterEtAl2010}} & MIS Guidance                            & Video             & Not Mentioned                      & Optical Marker    \\ 
                \textbf{Breton \textit{et al} (2010) \cite{Breton-LopezQueroBotellaEtAl2010}}  & Treatment of Cockroach Phobia           & Video             & HMD by 5DT (Model Unknow)          & Marker            \\ 
                \textbf{Alamri \textit{et al} (2010) \cite{AlamriChaElSaddik2010}}           & Poststroke-Patient Rehabilitation         & Video             & VUNIX iWear VR920                  & Marker            \\ \hline
                \end{tabular}

                 \begin{tablenotes}
                    \item[*] Kanade-Lucas-Tomasi algorithm \cite{TomasiKanade1991}.
                    \item[**] Iterative Closest Point algorithm \cite{BestMcKay1992}.
                \end{tablenotes}
            \end{threeparttable}
            \end{table*}

An alternative solution is to make use of projectors, half-silvered mirrors, or screens to augment information directly onto a physical space without the need to carry or wear any additional display devices.This is referred to as Spatial Augmented Reality (SAR) ~\cite{RaskarWelchFuchs1998}. By augmenting information in an open space, SAR enables sharing and collaboration. Video see-through SAR has been used for MIS guidance, augmenting the video output from an endoscope. Less common is optical see-through SAR but examples using semi transparent mirrors for surgical guidance have been built (e.g. ~\cite{LiaoInomataSakumaEtAl2010} ~\cite{WangSuenagaHoshiEtAl2014}), and also a magic mirror system for anatomy education ~\cite{MengFallavollitaBlumEtAl2013}. Another SAR approach is to insert a beam splitter into a surgical microscope to allow users to see the microscope view with an augmented image from a pico-projector ~\cite{ShiBeckerRiviere2012}. Finally, direct augmentation SAR will employ a projector or laser transmitter to project images directly onto the physical objects' surface. For example, the Spatial Augmented Reality on Person (SARP) system ~\cite{JohnsonSun2013} projects anatomical structures directly onto a human subject.   

Whatever display technology is used, particularly if it is a monocular display, a problem for MR is to provide an accurate perception of depth for the augmented information. Depth perception can significantly affect surgical performance \cite{Honeck2012}. In MIS, the problem is further compounded by large changes in luminance and the motion of the endoscope. Stereo endoscopes can be used to address this problem and they are available commercially. 3D depth information can then be recovered using a disparity map obtained from rectified stereo images during surgery  \cite{Stoyanov2005}. Surgeons have to wear 3D glasses, HMDs, or use binocular viewing interfaces in order to observe the augmented video stream.

\subsection{Registration} \label{registration}
Once the location for the augmented content has been determined, then this content (often computer-generated graphics) is overlayed or registered into the real word scene. Registration techniques usually involve an optimization step (energy function) to minimize the difference between virtual objects and real objects. For example, using the 3D to 3D Iterative Closest Point (ICP) \cite{BestMcKay1992} technique, or other 2D to 3D algorithms \cite{MarkeljTomazevicLikarEtAl2012}. The latter is also effective for registering preoperative 3D data such as CT and MRI images with intra-operative 2D data such as ultrasound, projective X-ray (fluoroscopy), CT-fluoroscopy, as well as optical images. These methods usually involve marker-based (external fiducial markers), feature-based (internal anatomical landmarks) or intensity-based methods that find a geometric transformation that brings the projection of a 3D image into the best possible spatial correspondence with the 2D images by optimizing a registration criterion. 
 
     	\begin{figure}[!htb]
    		\centering
    		\includegraphics[width=3.3in]{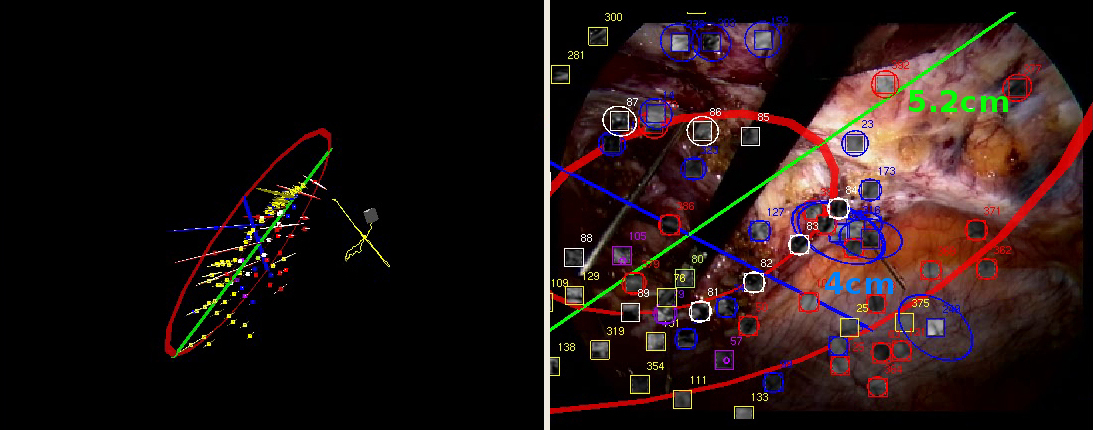}
        	\caption{The monocular SLAM system used in MIS \cite{GrasaBernalCasadoEtAl2014}. Left: Camera trajectory, 3D map and ellipses in 3D; Right: SLAM AR measurement, Map and ellipses over a sequence frame. Image courtesy of {\'{O}}scar G. Grasa, Ernesto Bernal, Santiago Casado, Ismael Gil and J. M. M. Montiel.}
        	\label{SLAM_fig}
    	\end{figure}
    	
\begin{figure}[!htb]
	\centering
	\includegraphics[width=3.3in]{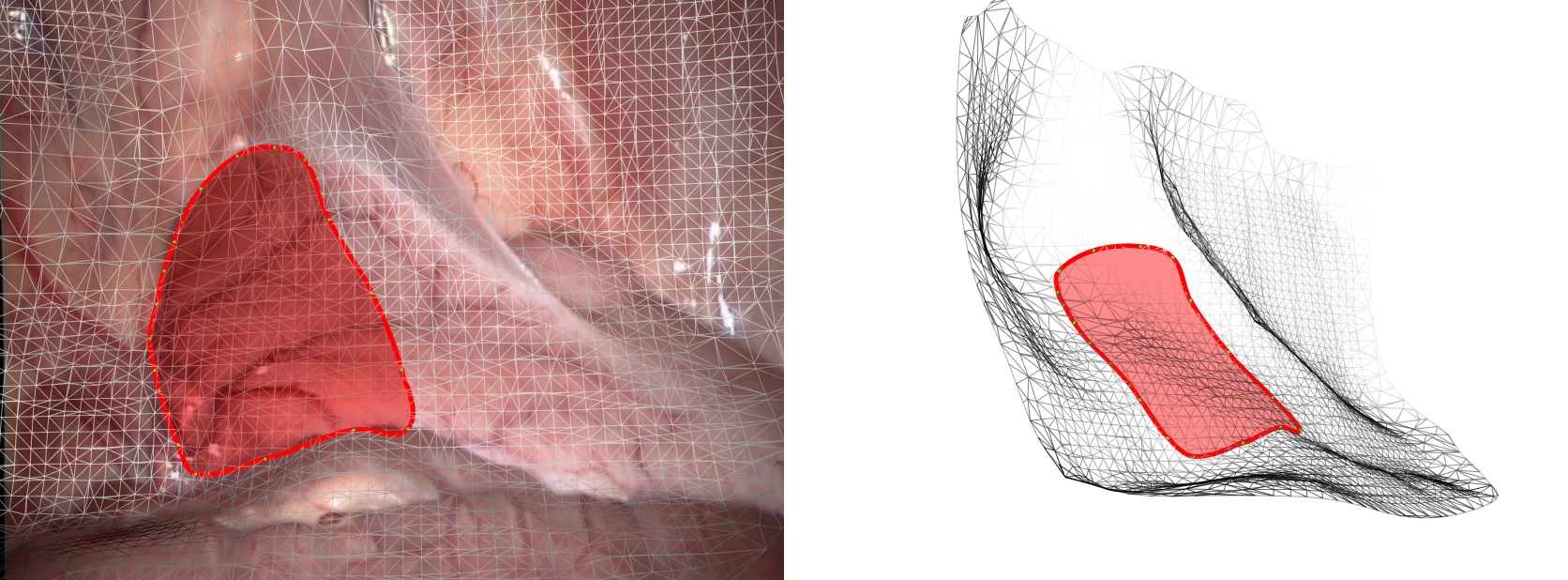}
	\caption{Geometry information can be acquired by stereo vision, providing geometry-aware AR environment in MIS enables interactive AR application such as intra-operative area highlighting and measurement. Note that the highlighting area (labeled in red) accurately follow alone the curve surface. ~\cite{ChenHTL2017}.}
	\label{mis_fig}
\end{figure}

Registration of virtual anatomical structures within MIS video is a much discussed topic ~\cite{LamataFreudenthalCanoEtAl2010} ~\cite{MarescauxDiana2015} ~\cite{MirotaIshiiHager2011} ~\cite{NicolauSolerMutterEtAl2011} ~\cite{DePaolisAloisio2010}. However, due to the problems of a limited field of vision (FOV), organ deformation, occlusion and no marker-based tracking possible, registration in MIS is still an unsolved problem. One possible approach is to use the Simultaneous Localisation And Mapping (SLAM) algorithm ~\cite{DissanayakeNewmanClarkEtAl2001} that was originally developed for autonomous robot navigation in an unknown space. AR has a very similar challenge as with robot navigation i.e. both need to get a map of the surrounding environment and locate the current position and pose of cameras ~\cite{CastleKleinMurray2008}. However, applying SLAM on single hand-held camera (such as an endoscopy camera) is more complicated than with robot navigation as a robot is usually equipped with odometry tools and will move more steadily and slowly than an endoscopy camera ~\cite{KleinMurray2007}. Grasa \textit{et al.} ~\cite{GrasaCiveraGuemesEtAl2009} proposed a monocular SLAM 3D model where they created a sparse abdominal cavity 3D map, and the motion of the endoscope was computed in real-time. This work was later improved to deal with the high outlier rate that occurs in real time and also to reduce computational complexity ~\cite{GrasaCiveraMontiel2011}, refer to Figure \ref{SLAM_fig}. A Motion Compensated SLAM (MC-SLAM) framework for image guided MIS has also been proposed ~\cite{MountneyYang2010}, which predicted not only camera motions, but also employed a high-level model for compensating periodic organ motion. This enabled estimation and compensation of tissue motion even when outside of the camera's FOV. Based on this work, Mountney and Yang \cite{MountneyFallertNicolauEtAl2014} implemented an AR framework for soft tissue surgery that used intra-operative cone beam CT and fluoroscopy as bridging modalities to register pre-operative CT images to stereo laparoscopic images through non-rigid biomechanically driven registration. In this way, manual alignment or fiducial markers were not required and tissue deformation caused by insufflation and respiration during MIS was taken into account. Chen \textit{et al.} ~\cite{ChenHTL2017} presented a geometry-aware AR environment in MIS by combining SLAM algorithm with dense stereo reconstruction to provide a global surface mesh for intra-operative AR interactions such as area highlighting and measurement. (see Figure \ref{mis_fig}).

\subsection{Tracking}
The tracking of real objects in a scene is an essential component of MR. However, occlusions (from instruments, smoke, blood), organ deformations (respiration, heartbeat) ~\cite{Puerto-SouzaMariottini2013} and the lack of texture (smooth regions and reflection of tissues) are all specific challenges for the medical domain.  

\begin{figure}[!htb]
        		\centering
        		\includegraphics[width=3.3in]{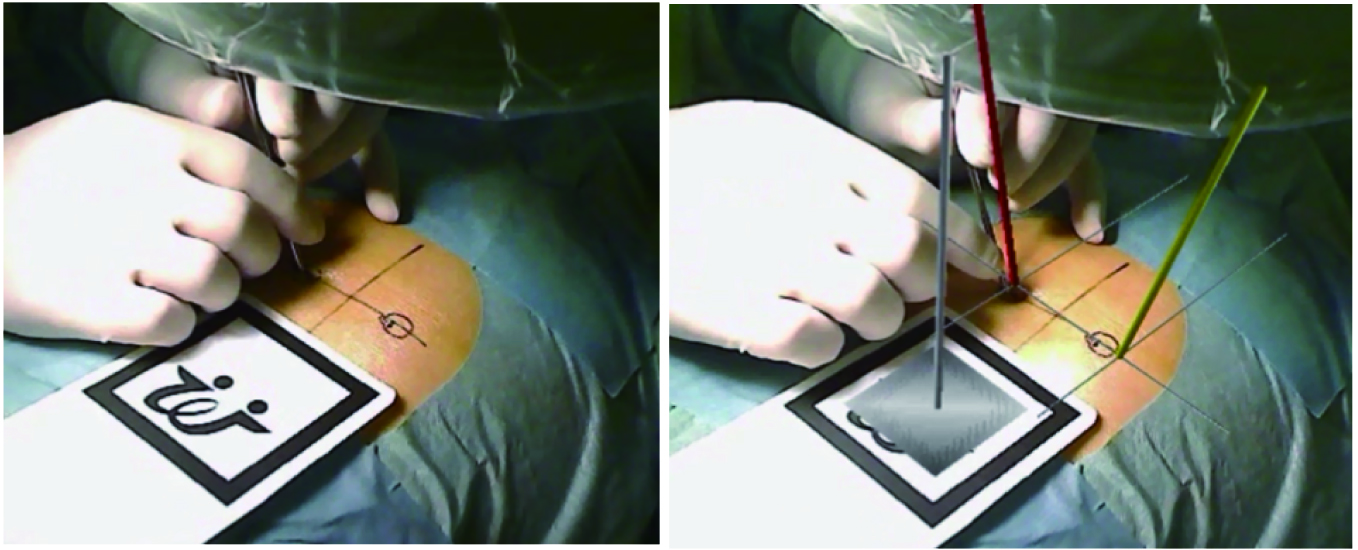}
            	\caption{A marker-based AR 3D guidance system for percutaneous vertebroplasty; the augmented red line and yellow-green line indicate the ideal insertion point and needle trajectory. Image courtesy of Yuichiro Abe, Shigenobu Sato, Koji Kato, Takahiko Hyakumachi, Yasushi Yanagibashi, Manabu Ito and Kuniyoshi Abumi \cite{AbeSatoKatoEtAl2013}.}
            	\label{tracking_fig1}
        	\end{figure}

Optical tracking, magnetic tracking, and the use of planar markers (with patterns or bar codes) have all been used in medical applications of MR. Some optical markers are specially made with iodine and gadolinium elements so that they can display high intensity in both X-ray/CT images and MR images. However, an optical tracking system also requires a free line-of-sight between the optical marker and the camera and this technology is impossible to use for MIS when the camera is inside the patient's body. Magnetic tracking in medical applications also lacks robustness due to interference caused by diagnostic devices or other ferromagnetic objects. Planar markers have been the most popular approach for tracking in AR to date and have been used successfully in medical application, for example, refer to Figure \ref{tracking_fig1}. However, care must be taken as planar markers are also prone to occlusion and have a limited detection range and orientation.
   
\begin{figure}[!htb]
        		\centering
        		\includegraphics[width=3.3in]{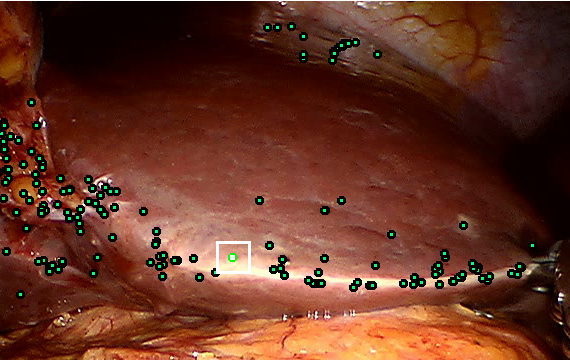}
            	\caption{The Speed-up Robust Feature (SURF) descriptor applied to endoscopic video of a liver. Image courtesy of Rosalie Plantef{\'{e}}ve, Igor Peterlik, Nazim Haouchine, St{\'{e}}phane Cotin \cite{PlantefevePeterlikHaouchineEtAl2016}.}
            	\label{tracking_fig2}
        	\end{figure}

 Markerless tracking is an alternative approach that utilizes the real-world scene and employs computer vision algorithms to extract image features as markers. Figure \ref{tracking_fig2} depicts one example from an endoscopic video of a liver. The quality of markerless tracking is dependent on lighting conditions, view angle and image distortion, as well as the robustness of the computer vision algorithm used. The performance of well known feature detection descriptors used in computer vision was evaluated in the context of tracking deformable soft tissue during MIS ~\cite{MountneyLoThiemjarusEtAl2007}.The authors present a probabilistic framework to combine multiple descriptors, which could reliably match significantly more features (even in the presence of large tissue deformation) than by using individual descriptors. 
 
 If images are acquired in a set of time steps, it is also possible to use optical flow to compute the camera motion and track feature points \cite{MirotaIshiiHager2011}. Optical flow is defined as a distribution of apparent velocities of brightness patterns in an image and can be used to track the movement of each pixel based on changes of brightness/light. Some recent work in medical MR ~\cite{HaouchineCotinPeterlikEtAl2015} ~\cite{Stoyanov2012} has been to combine computationally expensive feature tracking with light-weight optical flow tracking to overcome performance issues. If a stereoscopic display is being used then there are techniques developed to generate more robust and precise tracking, such as the real-time visual odometry system using dense quadrifocal tracking ~\cite{ChangHandaDavisonEtAl2014}.
In general, however, the accurate tracking of tissue surfaces in real time and realistic modelling of soft tissue deformation ~\cite{TangW14} remains an active research area.

\section{Research Challenges}
The feasibility of using MR has been demonstrated by an ever increasing number of research projects. This section summarises the ongoing challenges that are currently preventing MR becoming an accepted tool in every day clinical use. 

\subsection{Enabling Technologies}
Real time performance, high precision and minimum latency are pre-requisites in most medical applications. A typical MR system consists of multiple modules working together (image capture module, image detection and tracking module, content rendering module, image fusion module, display module, etc.) each of which has its own computational demands and each component can contribute to latency. Improvements to hardware and software continue to address these technology challenges. New generation hardware devices such as the HoloLens, Magic Leap, and next generation Google Glass will encourage further investigations and identify new problems. The FOV of the HoloLens is already being considered far too narrow for many applications. 

Many clinical procedures could benefit greatly from MR if patient specific data can be accurately delivered to the clinician to help guide the procedure. In such a system, if the augmented content were superimposed in the wrong position then the clinician could be misled and cause a serious medical accident. Many researchers are obtaining accuracy to within a few millimetres and this may be sufficient for some procedures and applications (an anatomy education tool, for example). Other procedures will need sub-millimetre accuracy. Automatic setup and calibration then becomes more critical. It remains challenging to find a balance between speed and accuracy as they are both very important in medical applications of MR. 

Accurate patient specific data modelling is also required to provide fully detailed information. Offline high-fidelity image capture and 3D reconstruction can provide some of the patient specific data needed for augmentation, but real-time high-fidelity online model reconstruction is also needed, for example, to handle tissue deformation. There is an urgent need for more research into real time high-fidelity 3D model reconstruction using online modalities such as video from endoscopic cameras. This will reduce the disparity between offline and real-time reconstruction performance capture. One method to address this challenge is to develop methods that target the existing low-resolution real-time tracker methods by adding local information. Developing a large database of 3D patient models is also a step towards bridging the gap. Artificial Intelligence algorithms could be used to learn the relationship between the online and offline image details to reconstruct the 3D information from the endoscopic camera. The goal is to train a model using data acquired by the high-resolution offline modality, which can then be used to predict the 3D model details given by online image capture. Breakthroughs in this area will provide both robustness and flexibility, as the training can be performed offline and then applied to any new patient.  

\subsection{MR for MIS}
 Currently a MIS surgeon has to continually look away from the patient to see the video feed on a nearby monitor. They are also hampered by occlusions from instruments, smoke and blood, as well as the lack of depth perception on video from the endoscope camera, and the indirect sense of touch received via the instruments being used. 

The smooth surface and reflection of tissues and the lack of texture make it hard to extract valid and robust feature points for tracking. The endoscope light source and its movement, as well as organ deformations caused by respiration and heartbeat, are changing the feature of each key point over time and destroying the feature matching processes. A general model for tracking cannot be used as the shape and texture information of an organ can vary from person to person, also, the limited FOV of endoscopic images restricts the use of model based tracking and registration methods. The only information that could be used for tracking is texture features. However, it is impossible to take a picture of the tracking target (such as a liver) before the operation, as the tracking targets (organs) are usually located inside human body. Some approaches \cite{MarkeljTomazevicLikarEtAl2012} \cite{ChenWangFallavollitaEtAl2013} have shown that it is possible to use pre-operative CT/MRI images to perform 3D to 2D registration to intra-operative X-ray fluoroscopy videos. However, these methods need to transfer the 3D CT/MRI image iteratively to find a best transformation, which can cost much precious time. In addition, patients' abdominal cavities are inflated with carbon dioxide gas to create the pneumoperitoneum, which will deform the original shape of organ and making it difficult for the deformable registration. Nevertheless, these methods demonstrate that matching features of 3D CT/MRI images with endoscopy video may be possible but still under great challenges. 

The latest MR systems in many application domains are using SLAM techniques and this approach is the current gold standard. However, real-time SLAM performance needs 3D points from a rigid scene to estimate the camera motion from the image sequences it is producing. Such a rigid scene is not possible in the MIS scenario. Research is therefore needed on how to cope with the tracking and mapping of soft tissue deformations and dealing with sudden camera motions. 

 Although stereo endoscopes are available, the majority of MIS procedures still use monoscopic devices and depth perception  remains a challenge. Recent work by Chen \textit{et al.} ~\cite{Chen2017} attempts to tackle this problem by providing depth cues for a monocular endoscope using the SLAM algorithm but more work is required, particularly when large tissue deformations occur. 

\subsection{Medical HCI}
HCI is an important aspect of MR. This encompasses speech and gesture recognition, which can be used to issue interaction commands for controlling the augmented scene. The use of haptics interfaces in MR are also starting to emerge. All of these require a constant highly accurate system calibration, stability and low latency. One approach that explored the visuo-haptic setups in medical training \cite{Harders2009} took into account of visual sensory thresholds to allow 100ms latency and one pixel of registration accuracy. This approach was able to minimize time lagged movements resulting from different delays of the data streams. With the new generation of portable AR headsets, in which mobile computing is central to the entire system setup, real-time computation will continue to be a challenge to achieve a seamless Haptics-AR integration.  

\subsection{Computing Power}
The capability of mobile graphics, with a move towards wearable wireless devices, is likely to be indispensable to the future of MR in medicine. Limited computing power, memory storage and energy consumption, however, will continue to be bottlenecks for real time 3D mobile graphics even with the advanced GPUs and hardware in current mobile devices. For example, the theoretical performance of the next generation of mobile GPUs is estimated at 400-500 GFLOPS, but 50\% of this will be reduced by the power and heat limitations. Now a major requirement of future  mobile applications will be extremely high resolution displays (i.e. 4K+). For the realistic rendering, physics and haptics responses required by medical MR applications,  a minimum of 30 fps (5000-6000 GFLOPS) is the performance target. This represents a 20 times increase on what is available from the current state of the art technology. The future of MR in medicine may therefore rely on cloud operations and parallelisation on a massive scale. 

\subsection{Validation Studies}
Any use of technology within a medical scenario must be proven to work, particularly where patients are involved. Nearly all of the surgical applications cited in this paper are carrying out these studies on phantom models or cadavers, and reported results are promising. The step to use MR for surgical guidance on live patients has yet to be made. The same is true for using MR for other forms of patient treatment. One of the few commercial tools currently available in the clinic - VeinViewer Vision - has provided mixed results from patient validation studies. For rehabilitation applications, the current situation is different. MR has already been shown to work with patients on both physical and neurological tasks. Rehabilitation is a promising area for the adoption of MR and AR techniques in the near future.

 As well as the technical challenges that have been the focus of this paper, there are other factors that must be overcome, for example, the often reluctance of the medical profession to embrace changes in their field \cite{gillam2009healthcare}. There is much momentum starting to appear, however, and we predict that the medical domain is on the cusp of adopting MR technologies into everyday practice.


\bibliographystyle{unsrt}
\bibliography{./Bibliographies/bibtex,./Bibliographies/Classifications/surgery,./Bibliographies/Classifications/education,./Bibliographies/Classifications/rehab,./Bibliographies/Classifications/training,./Bibliographies/Classifications/treatment,./Bibliographies/Technologies/interaction,./Bibliographies/Technologies/mobile,./Bibliographies/Technologies/display,./Bibliographies/Technologies/registration,./Bibliographies/Technologies/tracking,./Bibliographies/Challenge}
\end{document}